\def\year{2020}\relax
\documentclass[letterpaper]{article} % DO NOT CHANGE THIS
\usepackage{aaai20}  % DO NOT CHANGE THIS
\usepackage{times}  % DO NOT CHANGE THIS
\usepackage{helvet} % DO NOT CHANGE THIS
\usepackage{courier}  % DO NOT CHANGE THIS
\usepackage[hyphens]{url}  % DO NOT CHANGE THIS
\usepackage{graphicx} % DO NOT CHANGE THIS
\usepackage{graphicx}  % DO NOT CHANGE THIS
\title{Fast and Deep Graph Neural Networks}
\author{Claudio Gallicchio and Alessio Micheli\\
Department of Computer Science, University of Pisa\\
Largo B. Pontecorvo, 3\\
56127 Pisa, Italy\\
\{gallicch, micheli\}@di.unipi.it
}

\usepackage{amsthm}
\usepackage{url}
\usepackage{amsfonts}
\usepackage{booktabs}

\newtheorem{Definition}{Definition}
\newtheorem{Theorem}{Theorem}

\newcommand{\R}{\mathbb{R}}
\newcommand{\N}{\mathcal{N}}
\newcommand{\model}{FDGNN }
\newcommand{\A}{\mathbf{A}}
\renewcommand{\l}{\mathbf{l}}
\renewcommand{\u}{\mathbf{u}}
\newcommand{\x}{\mathbf{x}}
\newcommand{\y}{\mathbf{y}}
\newcommand{\W}{\mathbf{W}}
\newcommand{\Wi}{\W_I}
\newcommand{\Wh}{\W_H}
\newcommand{\Wy}{\W_Y}
\newcommand{\Wphi}{\W_\varphi}
\newcommand{\U}{\mathbf{U}}
\newcommand{\X}{\mathbf{X}}
\newcommand{\Z}{\mathbf{Z}}
\begin{document}

\maketitle

\begin{abstract}
We address the efficiency issue for the construction of a deep graph neural network (GNN). The approach exploits the idea of representing each input graph as a fixed point of a dynamical system (implemented through a recurrent neural network), and leverages a deep architectural organization of the recurrent units. Efficiency is gained by many aspects, including the use of small and very sparse networks, where the weights of the recurrent units are left untrained under the stability condition introduced in this work. This can be viewed as a way to study the intrinsic power of the architecture of a deep GNN, and also to provide insights for the set-up of more complex fully-trained models. Through experimental results, we show that even without training of the recurrent connections, the architecture of small deep GNN is surprisingly able to achieve or improve the state-of-the-art performance on a significant set of tasks in the field of graphs classification.
\end{abstract}

\section{Introduction}

Graphs  are relevant data structures that provide an useful abstraction for  many kind of  real data, ranging from molecular data to social and biological networks (just to mention the most noteworthy cases), and for all the cases characterized by data with relationships. The direct treatment of this kind of data allows  a Machine Learning (ML) system to consider the vector patterns, which characterize the standard flat domain, and the relationships among them, respecting in such a way the inherent nature of the underlying structured  domain.

It is not surprising, then, that there is a long tradition of studies for the processing of structured data in ML, starting for the Neural Networks (NN) since the ’90s, up to current increasing interest in the field of deep learning for graphs.  
Among the first NN approaches we can mention the Recursive Neural Network (RecNN) models for tree structured data 
in \cite{Sperduti1997,Frasconi1998}, and more recently in \cite{socher2011parsing,Socher2013recursive}, 
which have been progressively extended to directed acyclic graph \cite{Micheli2004contextual}. Such approaches provided a neural implementation of a state transition system traversing the input structures in order to make the embedding and subsequently the classification of the input hierarchical data. The main issue in extending such approaches to general graphs (cyclic/acyclic, directed/undirected) was the processing of cycles due to the mutual dependencies occurring among the state variables definitions represented in the neural recursive units.
The earliest approaches for graphs were the Graph Neural Network NN (GNN)\cite{scarselli2008graph}  and the Neural Network for Graphs (NN4G) \cite{micheli2009neural}. 
The GNN model is based on a state transition system similar to the RecNN that allows 
cycles  in the  state computation,  whereas the  stability of the recursive encoding process is guaranteed by resorting to a 
contractive state dynamics (Banach theorem for fixed point), which in turn is obtained by imposing  constraints to the loss function (alternating learning and convergence of the recursive dynamical system). 
In this approach, the context of each vertex is formed through graph diffusion during the iteration to the convergence of state dynamics. Theoretical approximation capability and VC dimension of GNN have been recently studied \cite{scarselli2018vapnik}. 
\\
The NN4G exploits the idea to treat the mutual dependencies among state variables managing them (architecturally) through different layers.  
without the use of recursive units  (also providing an automatic {\it divide et impera} approach for the architectural design). 
Instead of iterating at the same layer, each vertex can take the context of the other vertices computed in the previous layers, accessing progressively  to the entire graph context. 
Such idea  for a compositional embedding of vertices have been successively exploited in many forms in the area of spatial approaches, or convolutional NN (CNN) for graphs, which all share the process of traversing the graphs by neural units
with weight sharing  
(i.e. for  CNN the weights  are constrained to the neighbor topology  of all graph vertices  instead of the 2D matrix),
and the construction through many layers moving to deep architectures \cite{micheli2009neural,zhang2018end,tran2018filter,atwood2016diffusion,niepert2016learning,xu2018powerful}.
There are many other approaches in the field, including spectral-based NN approaches 
\cite{defferrard2016convolutional} 
and kernel for graphs
\cite{shervashidze2009efficient,yanardag2015deep,vishwanathan2010graph,neumann2016propagation,shervashidze2011weisfeiler}.

Summarizing, and following the general trend, learning in structured domains has progressively moved from flat to more structured forms of data (sequences, trees and up to general graphs), while NN models have been extended from shallow to deep architectures, allowing a multi-level abstraction of the input representation.  Unfortunately, both aspects imply a 
high computational cost, highlighting the need to move toward deep {\it and} efficient approaches for graphs. \\
For the case of sequences and trees, the Reservoir Computing (RC) paradigm provides an approach for the efficient modeling of recurrent/recursive models based on the use of fixed (randomized) values of the recurrent weights under stability conditions of the dynamical system (Echo State Property - ESP) \cite{Jaeger2004,Gallicchio2013tree}. 
Advantages for a deep construction of RC models in the sequential domain have been analyzed under different points of view, including the richness of internal representations, the occurrence of multiple time-scales (following the architectural hierarchy) and the competitive empirical results, see, e.g., \cite{gallicchio2018design}. Extension of the RC  to graphs \cite{gallicchio2010graph}  follows the line of GNN, whereas the stability condition is inherited by the contractivity of  the reservoir dynamics (ESP). 
However, the extension of such approaches to multi-layer architectures is still unexplored for general graphs.

Following all these lines, the aim of this work is to provide an approach to graph classification,
called Fast and Deep Graph Neural Network (FDGNN), which
combines:
(i) the  capability of 
stable dynamic systems (in the class of GNN models)  for the \emph{graphs embedding}, (ii)  the potentiality of a \emph{deep} organization of the GNN architecture (providing hierarchical abstraction of the input structures through the  architectural layers), and (iii) the extreme \emph{efficiency} of a randomized neural network, i.e. a model without training of the recurrent units. \\
The central idea is to exploit the fixed point of the recursive/dynamical system to represent (or {\it embed}) the input graphs. In particular, once the states have been computed for  all graph vertices,
iterating the state transition function until convergence,  such information is exploited  for the graph embedding and then projected to the model output, which in turn is implemented as a standard layer of trained neural units. \\
The efficiency issue will be addressed also in terms of sparse connections and a relative small number of units compared to the typical setting  of RC models or  compared to the  number of free-parameters of fully-trained approaches.
It is worth to note that, since that free-parameters of the embedding part do not undergo a training process, the model also provides a tool to investigate the inherent (i.e. independent from learning) architectural effect of layering in GNN.   
Hence, such  investigation can also provide  insights  for the set-up  of  more complex  fully-trained  models.\\
The paper also includes a theoretical analysis of  the condition for the stability of the neural graph embedding in deep GNN architectures.
The experimental analysis includes a significant set of  well-known  benchmarks in the field of graph classification, hence enabling a thorough comparison with recent approaches at the state-of-the-art.

\section{Proposed Method}
\textbf{Preliminaries on graphs}. 
In this paper we deal with graph classification problems. 
A graph $G$ is represented as a tuple $G = (V_G, E_G)$, where 
$V_G$ is the set of vertices (or nodes) and $E \subseteq V_G \times V_G$ is the set of edges. 
The number of vertices in $G$ is denoted by $N_G$.
The connectivity among the vertices in a given graph $G$ is represented by the adjacency matrix $\A_G \in \R^{N_G \times \N_G}$, where $\A_G(i,j) = 1$ if there is an edge between vertex $i$ and vertex $j$, and $\A_G(i,j) = 0$ otherwise. 
Although the proposed \model approach can be used to process both directed and undirected graphs, in what follows we assume undirected graph structures, i.e., graphs for which the adjacency matrix is a symmetric matrix.
The set of vertices adjacent to $v \in V_G$ is the neighborhood of $v$, i.e.,  $\N(v) = \{v' \in V_G : (v,v') \in E_G \}$. 
We use $k$ to indicate the degree of a set of graphs under consideration, i.e. the maximum among the sizes of the neighborhoods of the vertices. 
Each vertex $v$ is featured by a label, denoted by $\l(v)$, 
and which we consider to lie in a real vector space. When considering input graphs, we use $I$ to denote the size of vertex labels, i.e., $\l(v) \in \R^{I}$.
In what follows, when the reference to the graph in question is unambiguous, we drop the subscript $G$ to ease the notation.

\noindent
\textbf{The neural encoding process. }
The proposed \model model is based on constructing progressively more abstract neural representation of input graphs by stacking successive layers of GNNs. Crucially, and differently from the original formulation by the proponents of GNN \cite{scarselli2008graph}, in \model the parameters of each layers (i.e., the weights pertaining to the hidden units) are left untrained after initialization under stability constraints.

\begin{figure}[thb]
\center
 \includegraphics[width=.95\columnwidth]{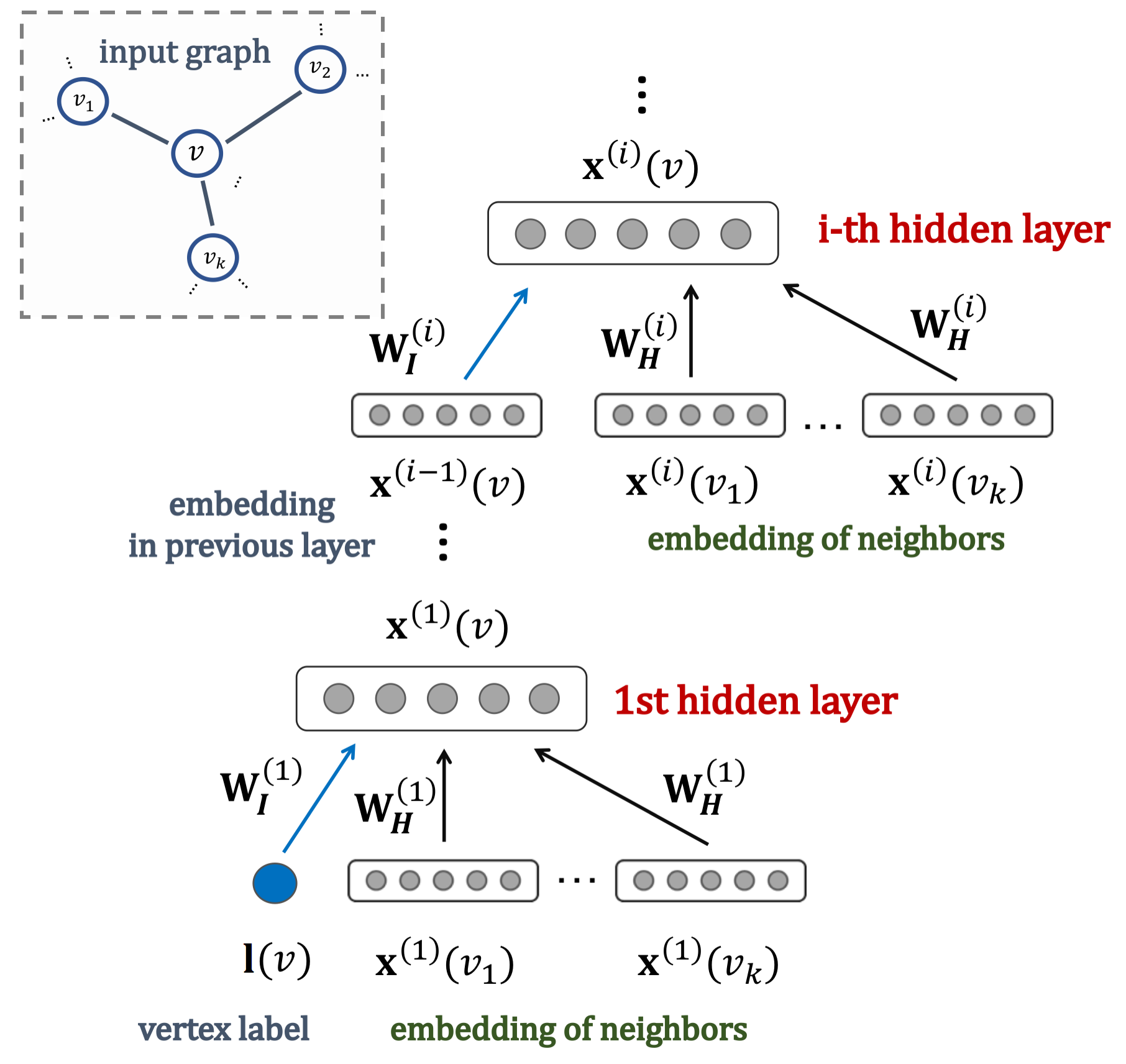}
    \caption{Deep encoding process implemented by FDGNN.} 
\label{fig.encoding}
\end{figure}

We use $L$ to denote the number of layers in the architecture. 
Then, for each layer $i = 1,\ldots, L$, the developed neural embedding (or state) for vertex $v$ is indicated by $\x^{(i)}(v)$. This is computed by the state transition function in layer $i$, which is implemented by neurons featured by hyperbolic tangent non-linearity, as follows:
\begin{equation}
\label{eq.state}
\x^{(i)}(v) = \tanh \big( \Wi^{(i)} \u^{(i)}(v) + \sum_{v'\in\N(v)} \Wh^{(i)} \x^{(i)}(v') \big),
\end{equation}
where, in relation to the given layer $i$, $\u^{(i)}(v) \in  \R^{U^{(i)}}$ is the input, $\Wi^{(i)} \in \R^{U^{(i)} \times H^{(i)}}$ is the input weight matrix that modulates the influence of the input on the representation, and $\Wh^{(i)} \in \R^{H^{(i)} \times H^{(i)}}$ is the recursive weight matrix that determines the influence of the neighbors representations in the embedding computed for $v$. Here, we use  $U^{(i)}$ and $H^{(i)}$, respectively, to denote the dimension of the input and embedding representations.
Moreover, here, and in the rest of the paper, references to bias terms are dropped in the equations for the ease of notation.
For all vertices $v$, the
embeddings $\x^{(i)}(v)$  are inizialized to zero values for all $i$.
Following the layered construction, for every vertex $v$ in the input graph, the input information for the first layer is given by the label attached to $v$, i.e., $\u^{(1)}(v) = \l(v)$ (and $U^{(1)} = I$). For successive layers $i>1$, the role of input information for vertex $v$ is played by the embedding developed for $v$ by the previous layer in the stack, i.e., $\u^{(i)}(v) = \x^{(i-1)}(v)$ (and $U^{(i)} = H^{(i-1)}$).
From the perspective of the analysis recently introduced in \cite{xu2018powerful},
FDGNN uses a neighborhood aggregation scheme that corresponds to a sum operator with untrained weights (summation term in eq.~\ref{eq.state}).
Figure~\ref{fig.encoding} 
shows the deep encoding process focused on a vertex $v$ of the input graph.

For any given layer $i \in \{1, \ldots, L\}$, the same state transition function in eq.~\ref{eq.state} is applied in correspondence of every vertex $v$ of an input graph.
The resulting operation can be conveniently and more compactly represented by resorting to a collective notation for the input and the embedding information, i.e., using $\U^{(i)} \in \R^{U^{(i)} \times N}$ and $\X^{(i)} \in \R^{H^{(i)} \times N}$ to column-wise collect respectively the input vectors $\u^{(i)}(v)$ and the embeddings $\x^{(i)}(v)$, for every $v \in V$.
In this way, the operation of the $i$-th hidden layer in the \model architecture can be described by means of a function
$F^{(i)}: \R^{U^{(i)} \times N} \times \R^{H^{(i)} \times N} \to \R^{H^{(i)} \times N}$, defined as follows:
\begin{equation}
\label{eq.F}
\X^{(i)} = F^{(i)}(\U^{(i)}, \X^{(i)}) = \tanh( \Wi^{(i)} \U^{(i)} + \Wh^{(i)}  \X^{(i)}  \A).
\end{equation}
In case of mutual dependencies among the vertices in the input graph (e.g., in the presence of cyclic substructures or of undirected connections) the application of eq.~\ref{eq.F} (or, equivalently of eq.~\ref{eq.state} in a vertex-wise fashion), might not admit a unique solution, i.e, a valid graph embedding representation.
It is then useful to study eq.~\ref{eq.F} as the iterated map that rules a discrete-time non-linear dynamical system, where $\X^{(i)}$ plays the role of state information, and $\U^{(i)}$ acts as the driving input.
To ensure the uniqueness of the developed neural representations, it is important that the system described by eq.~\ref{eq.F} is asymptotically stable, i.e., it converges to a unique fixed point upon repeated iteration.
The system dynamics are parametrized by the weight values in $ \Wi^{(i)}$ and $ \Wh^{(i)}$, which play a fundamental role in characterizing the resulting dynamical behavior. In the standard GNN formulation, the neural network alternates phases of state relaxation and learning of  $ \Wi^{(i)}$ and $ \Wh^{(i)}$ subject to contractive constraint employed through a penalizer term in the loss function for gradient descent \cite{scarselli2008graph}. 
In this paper, instead of employing a costly (and possibly long) constrained training process, we study stability conditions for the parameters in eq.~\ref{eq.F}, and initialize them accordingly. After initialization, the weights in $ \Wi^{(i)}$ and $ \Wh^{(i)}$ are left untrained, resulting in a graph encoding process that is extremely faster than that of classical GNNs, and is able to develop rich neural embeddings in a hierarchical fashion. The stability conditions for \model initialization are described in depth in Section~\ref{sec.dynamics}.
As a further characterizing aspect of the proposed approach, we make use of a \emph{sparse} design for both matrices $\Wh^{(i)}$ and $\Wi^{(i)}$. In particular, for every neuron in each hidden layer, we have only a constant number $C$ of connections from the previous layer (or from the input, for the first hidden layer), and of recurrent connections from the same layer. 
Note that the topology of the hidden layer architecture, i.e. the pattern of connectivity among the neurons, does not need to be related to the topology of the input graph. 
In particular, each neuron takes inputs from all the neighbour vertices embeddings (up to the degree $k$) from a number of neurons depending on $C$ (that controls the sparsity of connectivity in the NN layer). The topology of the input graph is hence preserved by the embedding process.

Based on the above described formulation, for any given input graph, the \model encoding process proceeds as follows. The state transition that rules the first hidden layer, i.e., $F^{(1)}$, is iterated until convergence to its fixed point, i.e., $\X^{(1)}$, driven by the vertices labels information as external input. Then, the same operation is repeated for the second layer, whose dynamics are ruled by $F^{(2)}$ and driven by $\X^{(1)}$, which now plays the role of input. This process goes on until the last layer $L$ is reached, and its state transition function $F^{(L)}$ converges to its fixed point $\X^{(L)}$. At each layer, the process of convergence to the fixed point is stopped whenever the distance between the states in successive iterations is below a threshold $\epsilon$, or a maximum number of iterations $\nu$ is reached.

\noindent\textbf{Output computation. }
For a given graph, the output of \model is computed by means of a simple readout layer that combines the neural representations developed by the last hidden layer of the architecture, as follows:
\begin{equation}
\label{eq.readout}
\y = \Wy \; \tanh \big(\Wphi \sum_{v \in V} \x^{(L)}(v) \big),
\end{equation}
where the argument of the hyperbolic tangent non-linearity expresses a combination of the embeddings computed in the last hidden layer for all the vertices in the graph, modulated by a projection matrix $\Wphi \in \R^{P \times H^{(i)}}$. Finally, $\Wy \in \R^{Y \times P}$ is the output weight matrix, where $\R^{Y}$ is the output space. 
In this work, elements in $\Wphi$ are randomly initialized from a uniform distribution and then are re-scaled and controlled to have unitary $L_2$-norm. The elements in $\Wy$ are the only free parameters of the model that undergo a training process, and are adjusted based on the training set. In the vein of RC approaches, training of the output weights is performed in closed form, exploiting the resulting convexity of the learning problem. This is performed by using Tikhonov regularization as described in \cite{Lukosevicius2009}. The readout operation implemented by eq.~\ref{eq.readout} can be seen as the application of a sum-pooling (or aggregation) operation, followed by a randomized non-linear basis expansion, and finally by a layer of trained linear neurons.

\subsection{Stability Conditions for Graph Embedding}
\label{sec.dynamics}
We study dynamical stability of the system implemented by each hidden layer of a GNN, as reported in eq.~\ref{eq.F}.
Here we focus our analysis on a generic layer $i$ of the architecture, and drop the $(i)$ superscript in the mathematical objects for the ease of notation. 
Accordingly, the dynamics under consideration are governed by the non-linear map $F$ that, given the input for the layer $\U$, and an initial state  $\X_0$ 
(set to zero values),
is iterated until convergence to the fixed-point solution, i.e. to the attractor of the dynamics. 

Here we use $F_{t}$ to denote the $t$-th iterate of $F$, and $\X_{t}$ to indicate the state after $t$ iterations, i.e.,
$\X_{t} = F(\U, \X_{t-1}) = F(\U, F(\U, \X_{t-2})) = \ldots = F(\U, F(\U, F(\ldots (F(\U,\X_0))\ldots)))$.
In the following, we assume that input and state spaces are compact sets, moreover we use $\| \cdot\|$ to indicate $L_2$-norm, and $\rho(\cdot)$ to indicate the spectral radius\footnote{
The largest abs value of the eigenvalues of its matrix argument.
}.
In order to develop a valid neural representation of input graphs, we require the system ruled by $F$ to be globally asymptotically stable, according to the following definition of \emph{Graph Embedding Stability} (GES).
\begin{Definition}[Graph Embedding Stability (GES)]
\label{def.ges}
For every input $\U$ to the current layer, and for every $\X_0, \Z_0$ initial states for the neural embeddings in the current layer, it results that:
\begin{equation}
\label{eq.ges}
\| F_t(\U,\X_0) - F_t(\U,\Z_0)\| \to 0 \quad as \;\; t \to \infty.
\end{equation}
\end{Definition}
GES in Definition~\ref{def.ges} essentially says that irrespective of the initial conditions, the developed neural encoding for the same input graph should be unique. Perturbations should vanish as the convergence process proceeds, and the resulting graph encoding is robust.
The property expressed by the GES can be also seen as a generalization of the ESP commonly imposed in the context of RC \cite{Jaeger2004}.
In the same vein as studies in RC, here we provide two conditions for the GES of the hidden layers in deep GNN, one sufficient, expressed by Theorem~\ref{th.sufficient}, and one necessary, expressed by Theorem~\ref{th.necessary}.
\begin{Theorem}[Sufficient condition for GES]
\label{th.sufficient}
For every input $\U$ to the current layer, if $\|\Wh\| \; k < 1$ then $F$ has dynamics that satisfy the GES.
\end{Theorem}
\begin{Theorem}[Necessary condition for GES]
\label{th.necessary}
Assume that a $k$-regular graph with null vertices labels is an admissible input for the system. If $F$ has dynamics that satisfy the GES then $\rho(\Wh) \; k < 1$.
\end{Theorem}
The proofs of both Theorems~\ref{th.sufficient} and  \ref{th.necessary} are reported in the Supplementary Material.

Theorems~\ref{th.sufficient} and \ref{th.necessary} provide a grounded way for stable initialization of hidden layers' in FDGNN.
In particular, the sufficient condition corresponds to a contractive setting of the neural dynamics for every possible input, and hence could be too restrictive in practical applications. Accordingly, we adopt the condition implied by Theorem~\ref{th.necessary} to initialize the weights in the hidden layers of the FDGNN, generalizing a common practice in the RC field \cite{Jaeger2004}.
As such, for every layer $i = 1, \ldots, L$, weight values in $\Wh^{(i)}$ are first randomly chosen from a uniform distribution in $[-1,1]$ and then re-scaled to have an effective spectral radius $\rho^{(i)} = \rho(\Wh^{(i)})\, k $, such that $\rho^{(i)} < 1 $. As pertains to the input matrices $\Wi^{(i)}$, their values are randomly sampled from a uniform distribution in $[- \omega^{(i)},\omega^{(i)}]$. The value of
$\omega^{(i)}$
acts as input scaling for $i=1$, and as inter-layer scaling for $i>1$.
We treat $\rho^{(i)}$  and $\omega^{(i)}$ as hyper-parameters.

\subsection{Computational Cost}
For any given input graph, the cost of the graph embedding process at layer $i$ is that of the iterated application of eq.~\ref{eq.F}.
Assuming $H$ hidden neurons, and exploiting the sparse connectivity in the hidden layer matrices, each iteration requires $\mathcal{O}((C+k+N) \, H)$. As such, the entire process of graph embedding for the whole architecture has a cost that is given by $\mathcal{O}(L\, \nu \, (C+k+N) \, H)$, which scales linearly with the number of neurons, with the number of layers, with the degree, and with the number of vertices (i.e., the size of the input graph).
Strikingly, the cost of the encoding  is the same for both training and test, as the internal weights do not undergo a training process and hence no additional cost for training is required. This leads to a clear advantage in comparison to fully trained NN models for graphs. 
Besides, the efficiency of the proposed approach is comparable to the Weisfeiler-Lehman graph kernel
\cite{shervashidze2011weisfeiler}, 
one of the most known and fastest kernels for graphs, whose  
cost scales linearly with the 
number of vertices and edges. 

The efficiency of FDGNN 
emerges also in the process of output computation, which is performed by a simple readout tool. 
This can be trained efficiently using direct methods, and generally is less expensive than training alternative readout implementations requiring gradient descent learning possibly through multiple layers (e.g., as in NNs for graphs), or support vector machines (e.g., as in kernel for graphs).

\section{Experiments}
We experimentally assess the proposed \model model on several benchmark datasets for graph classification, in comparison to state-of-the-art approaches from literature.

\noindent
\textbf{Datasets. }
We consider 9 graph classification benchmarks from the areas of cheminformatics and social network analysis. All the used datasets are publicly available online \cite{KKMMN2016}.
In the case of cheminformatics datasets, input graphs are used to represent chemical compounds, where vertices stand for atoms and are labeled by the atom type (represented by one-hot encoding), while edges between vertices represent bonds between the corresponding atoms. In particular, MUTAG \cite{debnath1991structure} is a collection of nitroaromatic compounds and the goal is to predict their mutagenicity on Salmonella typhimurium, PTC \cite{helma2001predictive} is a set of chemical compounds that are classified as carcinogenic or non-carcinogenic for male rats, COX2 \cite{sutherland2003spline} contains a set of cyclooxygenase-2 inhibitor molecules that are labeled as active or inactive, PROTEINS \cite{borgwardt2005protein} is a dataset of proteins that are classified as enzymes or non-enzymes, NCI1 \cite{wale2008comparison} contains anti-cancer screens for cell lung cancer. 
As regards the social network domain, we use IMDB-BINARY (IMDB-b), IMDB-MULTI (IMDB-m), REDDIT (binary) and COLLAB, proposed in \cite{yanardag2015deep}. IMDB-b and IMDB-m are movie collaboration datasets containing actor/actress ego-networks where the target class represent the movie genre, with 2 possible classes for IMDB-b (i.e., action or romance), and 3 classes for IMDB-m (i.e., comedy, romance, and sci-fi). REDDIT is a collection of graphs corresponding to online discussion threads, classified in 2 classes (i.e., question/answer or discussion community). Finally, COLLAB is a set of scientific collaborations, containing ego-networks of researchers classified based on the area of research (i.e., high energy, condensed matter or astro physics). Differently from the case of cheminformatics benchmarks, for the social network datasets no label is attached to the vertices in the graph. In this cases, we use a fixed (uni-dimensional) input label equal to $1$ for all the vertices. 
For binary classification tasks, a target value in $\{-1,1\}$ is considered. For multi-class classification tasks, the target label for each graph is a vector representing a binary $-1/+1$ one-hot encoding of the corresponding class.
A  summary of the datasets information is given in the Supplementary Material.
\\
\noindent
\textbf{Experimental settings. } 
We adopted a simple general setting, where all the hidden layers of the architecture in the graph embedding component shared the same values of the hyper-parameters, i.e., for $i\geq 1$ we set: number of neurons $H^{(i)} = H$ and effective spectral radius $\rho^{(i)} = \rho$; 
for $i>1$ we set the inter-layer scaling
$\omega^{(i)} = \omega$. 
In particular, we set the number of neurons in each hidden layer to $H = 50$ for 
all datasets except for NCI1 and COLLAB, for which we used $H = 500$. 
We implemented very sparse weight matrices with $C = 1$, i.e. every hidden layer neuron receives only 1 incoming connection from the previous layer (i.e., the input layer for neurons in the first hidden layer), and 1 incoming recurrent connection from a neuron in the same hidden layer. 
Values of $\rho$,  $\omega^{(1)}$ and $\omega$ were explored in the same range $(0,1)$. The above hyper-parameters of the graph embedding component were explored by random search, i.e., by randomly generating a number of $100$ configurations within the reported ranges. For every configuration, the value of the Tikhonov regularizer for training the readout was searched in a log-scale grid in the range $10^{-8} - 10^{3}$. Moreover, to account for randomization aspects, for every configuration we instantiated $20$ networks guesses, averaging the  error
on such guesses.
For the graph embedding convergence process
in each layer we used a threshold of $\epsilon = 10^{-3}$, and maximum number of iterations $\nu = 50$. The projection dimension for the readout was set to twice the number of neurons in the last hidden layer, i.e., $P = 2 \; H$. Accordingly, the total number of free parameters, i.e., the number of trainable weights for the learner, is as small as $1001$ for NCI1 and COLLAB, and $101$ for all the other datasets (note that there is an additional weight for the output bias). For all the datasets, we used the average value of $k$ in the dataset for network initialization.

For binary classification tasks, the output class for each graph was computed using the readout equation in eq.~\ref{eq.readout}, followed by the sign function for $-1 / +1$ discretization. In multi-class classification tasks, to each graph was assigned an output class in correspondence of the readout unit with the highest activation.
The performance on the graph classification tasks was assessed in terms of accuracy, and it was evaluated through a process of stratified 10-fold cross validation. For each fold, the \model hyper-parameter configuration was chosen by model selection, by means of a nested level of stratified 10-fold cross validation applied on the corresponding training samples. 

\subsection{Results}
The results 
achieved by FDGNN are reported in Table~\ref{tab.results}, where we show the test 
accuracy averaged over the outer 10 folds of the cross-validation (std are reported on the folds). 
We also give the performance achieved by \model settings constrained to a single hidden layer ($L = 1)$.
For comparison, in the same table we report the accuracy obtained by literature ML models for graphs. These includes a  variety of NN for graphs, i.e., 
GNN \cite{scarselli2008graph}, Relational Neural Networks (RelNN) \cite{blockeel2003aggregation}, Deep Graph Convolutional Neural Network (DGCNN) \cite{zhang2018end}, Parametric Graph Convolution DGCNN (PGC-DGCNN) \cite{tran2018filter}, Diffusion-Convolutional Neural Networks (DCNN) \cite{atwood2016diffusion}, PATCHY-SAN (PSCN) \cite{niepert2016learning}.
We also consider a number of relevant models from state-of-the-art graph kernels: Graphlet Kernel (GK) \cite{shervashidze2009efficient}, Deep GK (DGK) \cite{yanardag2015deep}, Random-walk Kernel (RW) \cite{vishwanathan2010graph}, Propagation Kernel (PK) \cite{neumann2016propagation}, and Weisfeiler-Lehman Kernel (WL) \cite{shervashidze2011weisfeiler}. Moreover, we consider two further recently introduced models: a recent hybrid CNN-kernel approach named Kernel Graph CNN (KGCNN) \cite{nikolentzos2018kernel}, and Contextual Graph Markov Model (CGMM) \cite{bacciu2018contextual}, which merges generative models and NN for graphs.
The performance 
in Table~\ref{tab.results} for the above mentioned approaches is quoted from the indicated reference (where the experimental setting for model selection was as rigorous as ours),
with the exception of PK and WL on COX2, which are quoted from \cite{neumann2016propagation}. When more than one configuration is reported in the literature reference, we quote the result of the one with the highest accuracy.

\begin{table*}[tbh]
  \caption{Test accuracy of FDGNN, compared to state-of-the-art results from literature. Performance of single hidden layer versions of \model ($L = 1$) are reported as well.  
  Results are averaged (and std are computed) on the outer 10 folds. Best results are highlighted in bold for every dataset.}
  \label{tab.results}
  \centering
  \begin{tabular}{llllll}
    \toprule
    
    & MUTAG & PTC & COX2 & PROTEINS & NCI1\\
    \midrule
    \model   & $\mathbf{88.51}_{\pm 3.77}$  & 
              $\mathbf{63.43}_{\pm 5.35}$ &
              $\mathbf{83.39}_{\pm 2.88}$ &  
              $\mathbf{76.77}_{\pm 2.86}$ &  
              $77.81_{\pm 1.62}$ \\
    FDGNN$_{(L = 1)}$  & ${87.38}_{\pm 6.55}$  & 
              $\mathbf{63.43}_{\pm 5.35}$ &
              ${82.41}_{\pm 2.67}$ &  
              $\mathbf{76.77}_{\pm 2.86}$ &  
              $77.11_{\pm 1.52}$ \\
    \midrule
    GNN \cite{uwents2011neural} & 
        ${80.49}_{\pm 0.81}$ & - & - & - & - \\
    RelNN \cite{uwents2011neural} & 
        ${87.77}_{\pm 2.48}$ & - & - & - & - \\
    DGCNN \cite{zhang2018end}
            & $85.83_{\pm 1.66}$ & $58.59_{\pm 2.47}$ &
              -                  & $75.54_{\pm 0.94}$ &
              $74.44_{\pm 0.47}$\\
    PGC-DGCNN \cite{tran2018filter}
            & $87.22_{\pm 1.43}$ & $61.06_{\pm 1.83}$ &
              -                  & $76.45_{\pm 1.02}$ &
              $76.13_{\pm 0.73}$\\
    DCNN \cite{tran2018filter}
            & - &
            - &
              -                  & 
              $61.29_{\pm 1.60}$ &
              $56.61_{\pm 1.04}$\\
    PSCN \cite{tran2018filter}
            & - &
            - &
              -                  & 
              $75.00_{\pm 2.51}$ &
              $76.34_{\pm 1.68}$\\  
    \\
    GK \cite{zhang2018end}
            & $81.39_{\pm 1.74}$ & $55.65_{\pm 0.46}$ &
              -                  & $71.39_{\pm 0.31}$ &
              $62.49_{\pm 0.27}$\\
    DGK \cite{yanardag2015deep}
            & $82.66_{\pm 1.45}$ & $57.32_{\pm 1.13}$ &
              -                  & $71.68_{\pm 0.50}$ &
              $62.48_{\pm 0.25}$\\
    RW \cite{zhang2018end}
            & $79.17_{\pm 2.07}$ & $55.91_{\pm 0.32}$ &
              -                  & $59.57_{\pm 0.09}$ &
              -\\
    PK \cite{zhang2018end}
            & $76.00_{\pm 2.69}$ & $59.50_{\pm 2.44}$ &
            $81.00_{\pm 0.20}$  & $73.68_{\pm 0.68}$ &
              $82.54_{\pm 0.47}$\\
    WL \cite{zhang2018end}
            & $84.11_{\pm 1.91}$ & $57.97_{\pm 2.49}$ &
            $83.20_{\pm 0.20}$  & $74.68_{\pm 0.49}$ &
              $\mathbf{84.46}_{\pm 0.45}$\\
    \\
    KCNN \cite{nikolentzos2018kernel}
            & - & $62.94_{\pm 1.69}$ &
              -                  & $75.76_{\pm 0.28}$ &
              $77.21_{\pm 0.22}$\\  
    CGMM \cite{bacciu2018contextual} &
            ${85.30}_{ }$ &
            - & - & - & - \\
    \bottomrule
  \end{tabular}
  \\
  \begin{tabular}{lllll}
    \toprule
    & IMDB-b & IMDB-m & REDDIT & COLLAB\\
    \midrule
    \model    & $\mathbf{72.36}_{\pm 3.63}$ & 
               $\mathbf{50.03}_{\pm 1.25}$ 
             & $\mathbf{89.48}_{\pm 1.00}$ & $74.44_{\pm 2.02}$\\
    FDGNN$_{(L = 1)}$    & 
    ${71.79}_{\pm 3.37}$ & 
               ${49.34}_{\pm 1.70}$ 
             & ${87.74}_{\pm 1.61}$ & $73.82_{\pm 2.32}$\\         
    \midrule
    DGCNN \cite{zhang2018end}
             & $70.03_{\pm 0.86}$ & $47.83_{\pm 0.85}$
             & - & $73.76_{\pm 0.49}$\\
    PGC-DGCNN \cite{tran2018filter}
            & $71.62_{\pm 1.22}$ & $47.25_{\pm 1.44}$ &
              -  & $\mathbf{75.00}_{\pm 0.58}$ \\
    PSCN \cite{tran2018filter}
            & $71.00_{\pm 2.29}$ & $45.23_{\pm 2.84}$ &
              -  & ${72.60}_{\pm 2.15}$ \\
    \\
    GK \cite{yanardag2015deep} &
            $65.87_{\pm 0.98}$ & $43.89_{\pm 0.38}$ &
            $77.34_{\pm 0.18}$ & $72.84_{\pm 0.56}$ \\
    DGK \cite{yanardag2015deep} &
            $66.96_{\pm 0.56}$ & $44.55_{\pm 0.52}$ &
            $78.04_{\pm 0.39}$ & $73.09_{\pm 0.25}$\\
    \\
    KCNN     \cite{nikolentzos2018kernel}
            & $71.45_{\pm 0.15}$ & $47.46_{\pm 0.21}$ &
              $81.85_{\pm 0.12}$ & $74.93_{\pm 0.14}$ \\

    \bottomrule
  \end{tabular}
\end{table*}

Results in Table~\ref{tab.results} indicate that \model outperforms the best literature results on 7 out of 9 benchmarks,
achieving state-of-the-art performance, and showing in many cases a clear improvement with respect to ML models in the area of NNs, kernel for graphs, and hybrid variants.
Moreover, even in the cases where \model does not achieve the top performance, its accuracy results to be the highest 
within the class of 
NN for graphs (for NCI1), or it is very close to the highest one (for COLLAB).

Results reported in Table~\ref{tab.results} are surprising, especially considering that they are obtained by a model in which the graph embedding process is not subject to training. Relevantly, the experimental results illustrated here highlight the ability of layered GNN architectures to construct neural embeddings for graph data that are rich in an \emph{intrinsic} way, i.e., even in the absence (or before) training of the recurrent connections.
Complementarily, our results point out the important role played by the aspect of \emph{stability} of the graph encoding process: as long as the layers of a deep GNN are able to implement such a process in a stable way, the neural representations developed in the hidden layers are \emph{per se} useful to effectively solve real-world problems in the area of graph classification.

Table~\ref{tab.results} also indicates the advantage of
depth in the architectural setup of  FDGNN. 
While the accuracy achieved by single hidden layer settings (i.e., for $L = 1$) is already comparable to literature results, deeper settings of \model consistently obtain a better performance (with the only exceptions of PTC and PROTEINS, where 1 hidden layer  configurations were selected by model selection).
As a further experimental reference, Table~\ref{tab.depth} shows the depth of the selected \model configuration for each dataset, averaged on the 10 folds. 
In most cases a number of layers of $\approx 3$ (and up to $5$) is selected, testifying the effectiveness of the multi-layered construction of the \model architecture. 

\begin{table}[htb]
  \caption{Depth of \model configurations selected by the nested cross validation process. Results are averaged (and std are computed) on the outer 10 folds.}
  \label{tab.depth}
  \centering
  \begin{tabular}{ll}
    \toprule
    Task & \# of layers\\
    \midrule
    MUTAG & $3.2_{\pm 1.0}$\\
    PTC & $1.0_{\pm 0.0}$\\
    COX2 & $2.7_{\pm 0.8}$\\
    PROTEINS & $1.0_{\pm 0.0}$\\
    NCI1 & $3.8_{\pm 0.6}$\\
    IMDB-b & $3.2_{\pm 1.2}$\\
    IMDB-m & $4.4_{\pm 0.8}$\\
    REDDIT & $3.0_{\pm 0.0}$\\
    COLLAB & $4.5_{\pm 0.8}$\\
    \bottomrule
  \end{tabular}
\end{table}

\begin{table}[htb]
  \caption{Running times required by \model (in single core mode, without GPU acceleration). Results are averaged (and std are computed) on the outer 10 folds.}
  \label{tab.times}
  \centering
  \begin{tabular}{lll}
    \toprule
    Task & Training & Test\\
    \midrule
    MUTAG & $0.56''_{\pm 0.33}$ & $0.06''_{\pm 0.04}$\\
    PTC & $0.16''_{\pm 0.03}$ & $0.02''_{\pm 0.00}$\\
    COX2 & $1.36''_{\pm 0.42}$ & $0.15''_{\pm 0.05}$\\
    PROTEINS &$2.16''_{\pm 0.47}$ & $0.24''_{\pm 0.04}$\\
    NCI1 & $2.00'_{\pm 0.45}$& $13.36''_{\pm 3.02}$\\
    IMDB-b & $7.46''_{\pm 3.14}$ & $0.83''_{\pm 0.35}$\\
    IMDB-m & $8.68''_{\pm 1.73}$ & $0.98''_{\pm 0.22}$\\
    REDDIT &  $2.47'_{\pm 0.01}$ & $16.49''_{\pm 0.28}$\\
    COLLAB & $22.86'_{\pm 4.70}$ & $2.54'_{\pm 0.52}$\\
    \bottomrule
  \end{tabular}
  \end{table}

The use of untrained recurrent connections makes the proposed \model strikingly efficient in applications. As evidence of this, in Table~\ref{tab.times} we report the running times for the training and test of the selected \model configurations, averaged on the 10 folds. The reported times are referred to a MATLAB implementation of FDGNN, running on a system with Intel Xeon Processor E5-263L v3 with 168 GB of RAM. Note that, although the \model model is amenable for parallelization, to provide an unbiased estimation of the required computation time, the reported running times were obtained using the system in single-core mode, and without any GPU acceleration.
Our code is made available 
%\footnote{Anonymized link: \url{https://bit.ly/32gmeSQ}}.
online\footnote{\url{https://bit.ly/32gmeSQ}}.
As it can be noticed, training and test times of FDGNN (even without parallelization) are generally very low, resulting from the combination of a number of factors: the sparse design of the hidden layers' matrices, the small number of neurons in each hidden layer, and the fact the internal weights are left untrained. Moreover, also the iterative embedding stabilization process taking place in each hidden layer is in practice not computationally intensive. Overall the approach results very fast, 
requiring a very small number of trainable weights (up to $1001$ in our experiments),
especially in comparison to literature models that entail much more complex architectural settings with multiple layers of fully end-to-end trained neurons, and possibly hundreds of thousands of trainable weights.

\begin{table}[htb]
  \caption{Comparison of training times required on MUTAG by FDGNN, GNN, GIN and WL. Results are averaged (and std are computed) on the outer 10 folds.}
  \label{tab.times_comparison}
  \centering
  \begin{tabular}{llll}
    \toprule
    FDGNN & GNN & GIN & WL\\
    \midrule
    $\mathbf{0.56''}_{\pm0.33}$ &
    $202.28''_{\pm166.87} $ &
    $499.24''_{\pm2.25} $
    &
    ${1.16''}_{\pm0.03}$\\
    \bottomrule
  \end{tabular}
  \end{table}
  
In order to give a more concrete idea on the efficiency of the proposed approach, in Table~\ref{tab.times_comparison} we compare the training times required by FDGNN on MUTAG with those required by GNN, GIN \cite{xu2018powerful}, and WL, as representatives respectively of dynamical NNs, recent convolutional NNs, and kernels models for graphs. In all cases, we used the code made available by the proponents, using the hyper-parameter settings given by the authors, and the same machine used for experiments on FDGNNs. Results in Table~\ref{tab.times_comparison} indicate that FGDNN 
requires the smallest training times, and results to be more than 300 times faster than GNN, almost 900 times faster than GIN, and comparable to WL (but still $\approx 2$ times faster). Relevantly, the FDGNN training speedup pairs the accuracy improvement already shown with respect to GNN and WL in Table~\ref{tab.results}, where accuracy of GIN from literature is not reported because obtained as validation accuracy of a 10-fold cross validation (while we are uniformly considering 
a 
more rigorous
nested cross validation with external test set). For completeness, the performance of GIN reported in \cite{xu2018powerful} ranges from $0.835$ to $0.9$ (depending on the variant). In similar experimental conditions, FDGNN achieves a test accuracy of $0.948$.

\section{Conclusions}
We have introduced FDGNN, a novel NN  model for fast learning in graph domains. The proposed approach showed that it is possible to combine the advantages of a deep architectural construction of GNN (in its ability to effectively process structured data in the form of general graphs), with the extreme efficiency of randomized NN, and in particular RC, methodologies. 
The randomized implementation allow us to implement untrained - but \emph{stable} - graph embedding layers, while through the deep architecture the model is able to build a progressively more  effective representation of the input graphs.
Despite the simplicity of the setup and the fast computation allowed by the model, the empirical accuracy of 
\model results to be very competitive with a large number of state-of-the-art CNN models and kernel methods for graphs.

\bibliography{references}
\bibliographystyle{aaai}

\newpage
\section*{Supplementary Material}
Here we report supplementary information, including the proofs of the theorems and additional information on the adopted datasets.
\\

\noindent
\textbf{Proof of Theorem 1}
We focus on the $i$-th hidden layer of the \model architecture. For any input $\U$ for the layer, and for any initial states $\X_0$ and $\Z_0$, we have that:
\begin{equation}
\label{eq.ps}
\begin{array}{l}
\| F_t(\U, \X_0) - F_t(\U, \Z_0)\| =\\
\| F(\U, \X_{t-1}) - F(\U, \Z_{t-1})\| = \\
\| \tanh(\Wi\,\U + \Wh \, \X_{t-1}\,\A) - \\
\quad\quad\quad\quad\quad\quad\quad\quad\quad\;
\tanh(\Wi\,\U + \Wh \, \Z_{t-1}\,\A) \| \leq\\
\| \Wh \| \, \| \A \| \| \X_{t-1} - \Z_{t-1} \| \leq\\
\ldots\\
(\| \Wh \| \, \| \A \|)^t \,  \| \X_0 - \Z_0 \| \leq\\
(\| \Wh \| \, k)^t \,  \| \X_0 - \Z_0 \|.
\end{array}
\end{equation}
Hence, if $\| \Wh \| \, k < 1$ then $\| F_t(\U, \X_0) - F_t(\U, \Z_0)\| $ converges to 0.
\qed
\\

\noindent
\textbf{Proof of Theorem 2}
We focus on the $i$-th hidden layer of the \model architecture.
For this case, we find it useful to build an equivalent dynamical system where the state is the row-wise concatenation of the state embeddings computed for each vertex in the input graph. We call this state $\tilde{\x}$. 
Using $\otimes$ to denote Kronecker product, the recurrent weight matrix of this system is given by $\tilde{\W}_H = \A \otimes \Wh$. I.e., $\tilde{\W}_H$ is
a block matrix where all the blocks are zero matrices, except for those corresponding to the positions of the non-zero entries in the adjacency matrix $\A$, which are set equal to $\Wh$.\\
We consider the linearized version of this new system around the zero state, and assume 
as input for the system 
a $k$-regular graph with null (i.e., zero) vertices labels. In this case, the linearized systems reads $\tilde{\x} = \tilde{\W}_H \tilde{\x}$. We know that if the condition $\rho(\tilde{\W}_H) < 1$ is violated, then the system is unstable around the zero state, and the GES property is not satisfied. Observing that the effective spectral radius $\rho(\tilde{\W}_H)$ is 
$\rho(\Wh)\, k$ (where $k$ is the degree), we can conclude that $\rho(\Wh)\, k < 1$ is a necessary condition for the GES.
\qed
\\

\noindent
\textbf{Datasets Information}
Tha major statistics on the datasets considered in our experiments are reported in Table~\ref{tab.datasets}.

\begin{table}[h]
\small
  \caption{Datasets information summary.}
  \label{tab.datasets}
  \centering
  \begin{tabular}{llll}
    \toprule
     Dataset & \# graphs & \# vertices tot (avg) & 
     \# classes\\
    \midrule
    MUTAG       & 188  & 3371   (17.9)  & 2 \\
    PTC         & 344  & 4915   (14.3)  & 2 \\
    COX2        & 467  & 19252  (41.2)  & 2 \\
    PROTEINS    & 1113 & 43471  (39.1)  & 2 \\
    NCI1        & 4110 & 122747 (29.9)  & 2 \\
    IMDB-b      & 1000 & 19773  (19.8)  & 2 \\
    IMDB-m      & 1500 & 19502  (13.0)  & 3 \\
    REDDIT      & 2000 & 859254 (429.6) & 2 \\
    COLLAB      & 5000 & 372474 (74.5)  & 3 \\
    \bottomrule
  \end{tabular}
  \end{table}

\end{document}